\documentclass{article}
\usepackage{spconf,amsmath,graphicx}
\usepackage{datetime}
\usepackage{listings}
\usepackage{graphicx}
\usepackage{float}
\usepackage{subfigure} 
\usepackage{cite}
\usepackage{caption} 
\usepackage{color}
\usepackage{amsthm}
\usepackage{bm}
\usepackage{mathrsfs}

\usepackage[noend]{algpseudocode}

\usepackage{algorithmicx,algorithm}
\usepackage[cmintegrals]{newtxmath}  
\usepackage{amsmath} 

\title{GAN-based Joint Activity Detection and Channel Estimation for Grant-Free Random Access}
\name{Shuang Liang$^{*}$$^{\dag}$$^{\ddag}$, Yinan Zou$^{*}$$^{\dag}$$^{\ddag}$, and Yong Zhou$^{*}$ 
\thanks{This work was supported by the National Natural Science Foundation of China (NSFC) under grant U20A20159.}}
\address{
	$^{*}$School of Information Science and Technology, ShanghaiTech University, Shanghai201210, China\\
	$^{\dag}$Shanghai Institute of Microsystem and Information Technology, Chinese Academy of Sciences, China\\
	$^{\ddag}$University of Chinese Academy of Sciences, Beijing 100049, China\\
E-mail:\{liangshuang, zouyn, zhouyong\}@shanghaitech.edu.cn}

\begin{document}
%
\maketitle
\begin{abstract}
 
\end{abstract}
Joint activity detection and channel estimation (JADCE) for grant-free random access is a critical issue that needs to be addressed to support massive connectivity in IoT networks. However, the existing model-free learning method can only achieve either activity detection or channel estimation, but not both. In this paper, we propose a novel model-free learning method based on generative adversarial network (GAN) to tackle the JADCE problem. 
We adopt the U-net architecture to build the generator rather than the standard GAN architecture, where a pre-estimated value that contains the activity information is adopted as input to the generator. By leveraging the properties of the pseudoinverse, the generator is refined by using an affine projection and a skip connection to ensure the output of the generator is consistent with the measurement. 
Moreover, we build a two-layer fully-connected neural network to design pilot matrix for reducing the impact of receiver noise.
Simulation results show that the proposed method outperforms the existing methods in high SNR regimes, as both data consistency projection and pilot matrix optimization improve the learning ability.
  \\

\section{Introduction}
Massive machine-type communications (mMTC), as an indispensable use case of the fifth generation (5G) wireless networks, is envisioned to provide massive connectivity for a large number of Internet of Things (IoT) devices\cite{massive_connectivity, Shi2022Algorithm}. However, the conventional grant-based random access scheme\cite{Zou2021Learning} incurs excessive signaling overhead, and there are not enough orthogonal signature sequences to be allocated to all IoT devices. To address these issues, the 3rd generation partnership project (3GPP) for 5G new radio (NR) proposes the grant-free random access\cite{grant_free}, where each device can directly send a unique non-orthogonal pilot sequence together with its data, without waiting for the permission from the base station (BS). Joint activity detection and channel estimation (JADCE)\cite{JADCE} is recognized as a critical  issue for realizing grant-free random access \cite{Mostafa2019}. 
\par Various methods have been proposed to tackle the JADCE problem and can generally be divided into two categories: optimization-based methods and learning-based methods. Despite the fact that the optimization-based methods, such as the
block coordinate descent algorithm for Group Lasso\cite{yuan2006model} and iterative shrinkage thresholding algorithm (ISTA)\cite{ista}, can solve the JADCE problem effectively, they require a large number of iterations.
In order to accelerate the convergence and improve the estimation performance, learning-based methods have recently been proposed to tackle the JADCE problem. Learning-based methods can be further divided into model-based methods and model-free methods. For the model-based methods, the authors in \cite{lista2} proposed a learned ISTA (LISTA). By integrating the unrolled deep neural network and ISTA, LISTA achieves a linear convergence rate. In addition, the authors in \cite{CY} proposed two model-based methods, which utilize data features to achieve better performance by combining the standard auto-encoder structure with an iterative optimization algorithm, such as Group Lasso and approximate message passing (AMP) method\cite{grant_free, JADCE}.
\par 
Compared to the model-based methods, the model-free methods have stronger adaptability and learning ability. The existing studies all adopt the generative network, such as Auto-Encoder (AE), Variational AE (VAE)\cite{VAE} and standard Generative Adversarial Network (GAN)\cite{gan}. In particular, the authors in \cite{AE} proposed a model-free method to jointly design the pilot matrix and detect active devices by using AE. The authors in \cite{ganhigh} used GAN to achieve high dimensional wireless channel estimation by optimizing the input to a deep generative network. However, these model-free methods can only achieve either activity detection or channel estimation, but not both.
In addition, the generative network in \cite{ganhigh} learns the mapping from the latent variable with a specific distribution to the wireless channel distribution. As the generative network is fixed and the latent variable is optimized according to the measurement error. 
This generative network is unable to learn the feature of active devices. In addition, the optimization about the latent variable cannot ensure that its distribution does not change, which may conflict with the mapping of the generator and make the latent variable falls into local optimum. By combining the advantages of \cite{AE} and \cite{ganhigh}, a pre-estimated value containing active device information can be adopted as input to the generative network to tackle the JADCE problem, which motivates this work. 
\par In this paper, we propose a novel model-free method to address the JADCE problem for grant-free random access in IoT networks. Instead of adopting the standard generative network architecture, we adopt the U-net\cite{unet,Unet1,Unet2} architecture to build generator which can be trained in an end-to-end manner based on a pre-estimated value.
According to the properties of pseudoinverse, we project the output of the U-net to the nullspace of the  pilot matrix to ensure the data consistency. Moreover, a two-layer fully-connected neural network was built to design the pilot matrix to reduce the impact of receiver noise. Simulation results show that the proposed method solves the problem of latent variable distribution variation and achieves a better performance than the existing methods for the JADCE problem in high signal-to-noise ratio (SNR) regimes.

\section{System Model}

Consider the grant-free uplink transmission of a single-cell IoT network that consists of $N$ single-antenna IoT devices and one $M$-antenna BS.
We denote $\mathcal{N}=\{1,\dots,N\}$ as the index set of IoT devices and assume that the transmission activity of each IoT device is independent in each transmission block. 
We denote $a_n$ as the indicator of the activity of device $n$, where $a_n=1$ if device $n$ is active and $a_n=0$ otherwise. 
We denote $\bm{y}(l)\in \mathbb{C}^M$ as the $l$-th symbol received by the BS, $\bm{h}_n\in \mathbb{C}^M$ as the channel coefficient vector of the link between IoT device $n$ and the BS, $s_n(l)\in \mathbb{C}$ as the $l$-th signature symbol transmitted by device $n$, and $\bm{z}(l)\in \mathbb{C}^M$ as the additive white Gaussian noise (AWGN) vector at the BS. 
By assuming that the transmissions of active devices are synchronized, $\bm{y}(l)\in \mathbb{C}^M$ can be written as $\bm{y}(l)=\sum_{n=1}^{N}a_n \bm{h}_n s_n(l)+\bm{z}(l),l=1,\dots,L$,
where $L$ denotes the length of the signature sequence.

\par For notational ease, we denote the aggregated received pilot signal matrix $\bm{Y}\in \mathbb{C}^{L\times M}$ with $\bm{Y}[l,:]=\bm{y}(l)$,
the channel matrix $\bm{H}\in \mathbb{C}^{N\times M}$ with $\bm{H}[n,:]=\bm{h}_n$,
the additive noise matrix $\bm{Z}\in \mathbb{C}^{L\times M}$ with $\bm{Z}[l,:]=\bm{z}(l)$,
and the pilot matrix $\bm{S}\in \mathbb{C}^{L\times N}$ with $\bm{S}[l,:]=[s_1(l),\dots,s_N(l)]$. We have
\begin{gather}
\bm{Y}=\bm{SX}+\bm{Z},
\end{gather} 
where matrix $\bm{X}=\bm{AH} \in C^{N\times M}$ with $\bm{A}=\mathrm{diag} \left( a_1,\dots,a_N\right)$ being the diagonal activity matrix. 
By considering the sporadic traffic, matrix $\bm{X}$ endows with a group sparse structure which means that all columns of matrix $\bm{X}$ have the same sparse structure. 
The goal of the JADCE problem is to detect the device activity matrix $\bm{A}$ and estimate the channel matrix $\bm{H}$ by recovering $\bm{X}$ from the noisy observation $\bm{Y}$.


\par As the standard neural networks cannot process complex-valued data, we reformulate (1) as 
\begin{equation}
\begin{split}
\tilde{\bm{Y}} &= \tilde{\bm{S}}\tilde{\bm{X}}+\tilde{\bm{Z}}\\
&=\left[\begin{array}{cc}
\mathcal{R} \{\bm{S}\}& -\mathcal{I}\{\bm{S}\}\\
\mathcal{I}\{\bm{S}\} & \mathcal{R} \{\bm{S}\}
\end{array}\right]
\left[\begin{array}{c}
\mathcal{R} \{\bm{X}\}\\
\mathcal{I} \{\bm{X}\}
\end{array}\right]
+\left[\begin{array}{c}
\mathcal{R} \{\bm{Z}\}\\
\mathcal{I} \{\bm{Z}\}
\end{array}\right]
\end{split}
\end{equation}
where $\mathcal{R}\{\cdot\}$ and $\mathcal{I}\{\cdot \}$ denote the real part and imaginary part of the input matrix, respectively.
\begin{figure*}[t]
	\centering
	\includegraphics[scale=0.5]{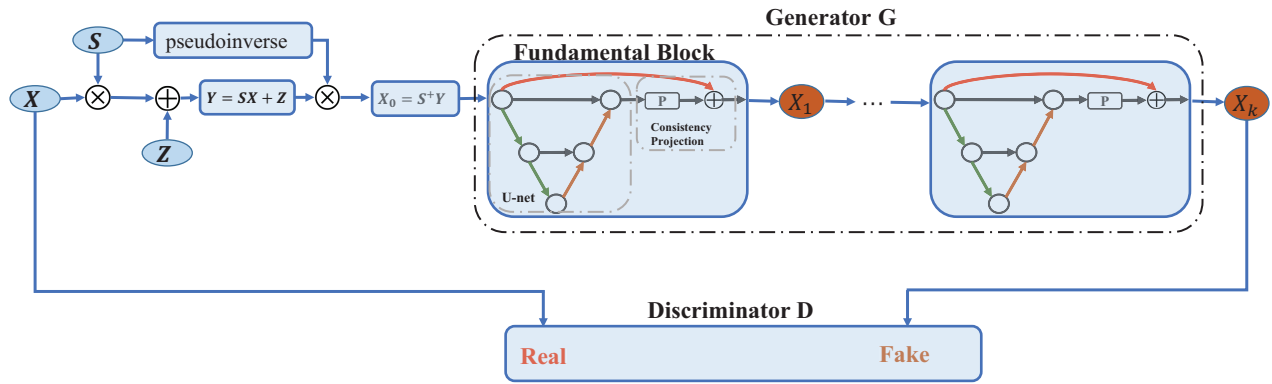}
	\caption{Proposed model-driven approach.}
	\label{fig:label}
	\vspace{-5mm}
\end{figure*}
\section{Algorithm Description}
\subsection{Overview of the Proposed Method }
\par Fig. 1 shows an overview of the proposed GAN-based method for the JADCE problem. For clarity, we represent a U-net with data consistency projection as a fundamental block. Our generator $G$ is made up of many fundamental blocks. With $\tilde{\bm{Y}}= \tilde{\bm{S}}\tilde{\bm{X}}+\tilde{\bm{Z}}$, we use $\tilde{\bm{X}}_0=\tilde{\bm{S}}^+\tilde{\bm{Y}}$ as an estimation of $\tilde{\bm{X}}$ to be the input of the generative network, where $(\cdot)^+$ is the pseudoinverse of a matrix. Inspired by the residual learning and the properties of the Moore–Penrose pseudoinverse, we build a shortcut connection from input to output in each fundamental block, which guarantees the data consistency. Our discriminator $D$ attempts to differentiate the real group sparse instances from the fake instances generated by $G$. The whole system trains $G$ and $D$ adversarially until the loss of each neural network converges.
\par 
Training the network in Fig. 1 is equivalent to playing a game with conflicting objectives between the generator $G$ and the discriminator $D$. The generator $G$ aims to map the estimation $\tilde{\bm{S}}^+\tilde{\bm{Y}}$ to the fake data that fools the discriminator $D$. The discriminator $D$ aims to distinguish between real data and data generated by the generator $G$. Various strategies have been devised to reach the balance. Motivated by LSGAN\cite{LSGAN} and WGAN\cite{wgan}, we adopt a mixture of WGAN and $l_2$ costs to train the generator. We then formulate the overall problem to jointly minimize the generator cost $
\mathbb{E}[||\tilde{\bm{X}}-G(\tilde{\bm{S}}^+\tilde{\bm{Y}})||_2]-\mathbb{E}[D(G(\tilde{\bm{S}}^+\tilde{\bm{Y}}))]
$ and the discriminator cost $
-\mathbb{E}[D(\tilde{\bm{X}})]+\mathbb{E}[D(G(\tilde{\bm{S}}^+\tilde{\bm{Y}}))].
$
\subsection{U-net}
The standard generative models, such as AE and VAE, map the input data to a low-dimensional latent variable to extract the main features of the input data.
They feed the latent variables into decoder to recover data. Hence, these methods have to find the corresponding latent variables to recover the input data. 
However, finding the corresponding latent variables is not necessary for the JADCE problem and the dimensional reduction inevitably leads to performance loss.
On the other hand, the standard GAN takes low-dimensional random latent variables that obey a specific distribution as input. 
A trained generator can generate many similar data to approximate the real data for different latent variables. 
The authors in \cite{AE} used $\tilde{\bm{Y}}$ as input to AE to achieve a good performance on activity detection, which verifies that $\tilde{\bm{Y}}$ contains the device activity information. The authors in \cite{ganhigh} used standard GAN to achieve a good performance for channel estimation. 
Due to the independence of activity among devices, the method in \cite{ganhigh} is unable to achieve activity detection.
\par By leveraging their advantages, we propose a novel GAN architecture which exploits the U-net for the generator.
Under this structure, we use $\tilde{\bm{S}}^+\tilde{\bm{Y}}$ as the input to the generator, which contains the device activity information, to tackle the JADCE problem.
By utilizing a shortcut connection of the U-net, it is not necessary to find the corresponding latent variables, which may introduce many additional iterations.
In addition, the shortcut connections also enhance the feature reuse to reduce information loss.

\subsection{Data Consistency}
For a linear equations $\tilde{\bm{Y}}=\tilde{\bm{S}}\tilde{\bm{X}}$, where $\tilde{\bm{S}}$ is an underdetermined matrix, $\tilde{\bm{X}}$ is variable and $\tilde{(\cdot)}$ is the real-valued form of matrix, all the solutions satisfy $\tilde{\bm{X}}=\tilde{\bm{S}}^+\tilde{\bm{Y}}+(\bm{I}-\tilde{\bm{S}}^+\tilde{\bm{S}})\tilde{\bm{W}}$, where $\bm{W}$ is an arbitrary matrix. To ensure the data consistency, we project the output of U-net onto the nullspace of $\tilde{\bm{S}}$ and add $\tilde{\bm{S}}^+\tilde{\bm{Y}}$ to arrive at $\tilde{\bm{X}}_1=\tilde{\bm{S}}^+\tilde{\bm{Y}}+(\bm{I}-\tilde{\bm{S}}^+\tilde{\bm{S}})U_1(\tilde{\bm{S}}^+\tilde{\bm{Y}})$ , where $U_k(\cdot)$ represents the output of the $k$-th U-net. Subsequently, we get $\tilde{\bm{X}}_2=\tilde{\bm{X}}_1+(\bm{I}-\tilde{\bm{S}}^+\tilde{\bm{S}})U_2(\tilde{\bm{X}}_1)$ which also satisfies data consistency. As a result, we can model it as a recurrent neural network (RNN). Moreover, we can express the unrolled RNN with $k$ iterations as 
\begin{gather}
\tilde{\bm{X}}_{k+1}=\tilde{\bm{X}}_k+(\bm{I}-\tilde{\bm{S}}^+\tilde{\bm{S}})U_k(\tilde{\bm{X}}_k),
\end{gather}
where $\tilde{\bm{X}}_{k}$ is the $k$-th output of fundamental block and $U_k$ is the $k$-th U-net of the fundamental block.  

\par To account for the noise $\bm{Z}$, we use $\tilde{\bm{X}}_{k+1}=\tilde{\bm{X}}_k+(\bm{I}-\tilde{\bm{S}}^+\tilde{\bm{S}})U_k(\tilde{\bm{X}}_k)-\tilde{\bm{S}}^+\tilde{\bm{Z}}$ to ensure the data consistency instead of (3). However, as the receiver noise is random, we cannot completely eliminate the effect of noise. In order to reduce the impact of noise, the pilot matrix $\tilde{\bm{S}}$ is designed to minimize $\mathbb{E}[||\tilde{\bm{S}}^+\tilde{\bm{Z}}||_F]$.
Motivated by \cite{AE}, we build a two-layer fully-connected neural network, which the input layer has $2N$ neurons and the output layer has $2L$ neurons. 
In addition, the weight of the connection from the $n$-th neuron in the input layer to the $l$-th
neuron in the output layer represents the $(l, n)$-th element of the pilot matrix $\tilde{\bm{S}}$. The  $\mathbb{E}[||\tilde{\bm{S}}^+\tilde{\bm{Z}}||_F]$ is the loss function. Given training dataset $\{\tilde{\bm{Z}_i}\}_{i=1}^n$ and a initial value of ${\tilde{\bm{S}}}$, we use gradient descent algorithm minimize the loss function until the convergence.

\subsection{Model Architecture}
In this subsection, we introduce the details of our neural network architecture. Firstly, we introduce the network architecture of generator. The generator is made up of multiple fundamental blocks, which include two components:  U-net block and projection block. By denoting the input of fundamental block as $\tilde{\bm{X}}_0$, we can get the output of U-net block $U(\tilde{\bm{X}}_0)$ and the output of projection block $\tilde{\bm{X}}_1  =\tilde{\bm{X}}_0+(\bm{I}-\tilde{\bm{S}}^+\tilde{\bm{S}})U(\tilde{\bm{X}}_0)$. The input of U-net block is a 3D tensor. 
From the dimensional perspective, the dimensions of the input data first reduce and then increase during the feedforward of the input data. In the process of reducing dimension, we first use 1D convolution with a kernel size of 3, stride size of 2 and padding size of 1, so that it performs down-sampling without a separate max-pooling layer. Then, we use 1D convolution with a kernel size of 3 and stride size of 1 and ReLU activation function to extract features, and repeat this process twice. In the process of increasing dimension, we use 1D convtranspose with a kernel size of 2 and stride size of 2. Then we stack the tensor with the tensor which has the same size in the processing of reducing dimension. Next, we use 1D convolution with a kernel size of 3, stride size of 1 and ReLU activation function to extract features, and repeat this process twice. 
\par Secondly, we design the network architecture of the discriminator as follows. We use 1D convolution with a kernel size of 3, stride size of 1 and ReLU activation function to extract features. Then we use 1D convolution with a kernel size of 3 and stride size of 2, so that it performs down-sampling. After several dimensional reduction operations, we use 1D convolution with a kernel size of 3 to get final output.
\section{Simulation Results}
\subsection{Data Generation and Training Strategy}
In this section, we present the simulation results of the proposed method, and compare the results with the classic optimization-based methods, ISTA\cite{ista} and Group Lasso\cite{yuan2006model}, and the state-of-the-art model-based method LISTA\cite{lista2}.
In the simulations, the signature sequence is generated according to the complex Gaussian distribution, i.e, $\bm{S}\sim \mathcal{CN}(0,1)$, and the channels suffer from independent Rayleigh fading, i.e., $\bm{H}\sim \mathcal{CN}(0,1)$. We set the total number of devices $N$, the length of signature sequence $L$, and the number of antennas at the BS $M$ to be 256, 128, and 8, respectively. Each entry of the activity sequence $\{a_1, ..., a_N \}$ follows the Bernoulli distribution with probability $p = 0.1$, i.e, $\mathcal{P}(a_n=1)=p$ and $\mathcal{P}(a_n=0)=1-p$. According to (2), we transform all the complex-valued matrix into real-valued matrix. And then, we obtain the data set $\{\tilde{\bm{X}}_i,\tilde{\bm{Y}}_i\}_{i=1}^{n}$, where $n$ represents the size of the data set.  We adopt normalized mean square error (NMSE) to evaluate the performance of those methods, defined as 
\begin{gather}
NMSE=10\log_{10}\Bigg(\frac{\mathbb{E}||{\tilde{\bm{X}}}-{\tilde{\bm{X}}^*}||_F^2}{\mathbb{E}{||\tilde{\bm{X}}^*||_F^2}}\Bigg),
\end{gather}
where $\tilde{\bm{X}}$ is the estimate solution and $\tilde{\bm{X}}^*$ is the ground truth. In the the training stage, we set the batch size to 64 and the initial learning rate $\eta_0$ to $5*10^{-4}$. In the test stage, we generate 1000 samples to test the proposed model.
\par We train the our proposed model by adopting the block-wise training strategy. To stabilize the training process, we add two decayed learning rates $\eta_1$ and $\eta_2$, i.e., $\eta_1=0.2\eta_0$ and $\eta_2=0.02\eta_0$. We train the generator $G$ block by block. The training process of each block is described as follows. First, we suppose that the previous blocks have been trained and train $i$-th block with learning rate $\eta_0$ until the convergence. And then, we train all blocks with learning rate $\eta_1$ and $\eta_2$ until the convergence. Finally, we add new block and repeat the process. The code is available at \\\texttt{https://github.com/deeeeeeplearning/JADCE}

\subsection{Experiment}
\begin{figure}[t]
	\centering
	\includegraphics[scale=0.35]{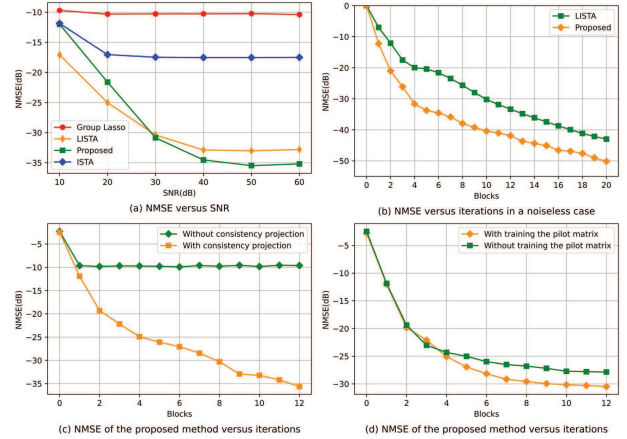}
	\caption{Performance comparison in terms of NMSE when $N$=256, $L$=128, $M$=8, and $p$=0.1.}
	\label{fig:label}
	\vspace{-5mm}
\end{figure}
In Fig. 2(a), we compare the performance of our proposed method with the baseline methods over different SNR. When SNR is less than 30 dB, the performance of our proposed method is better than ISTA and Group Lasso, but not better than LISTA. However, the performance of the proposed method improves rapidly with the increase of SNR. When SNR is greater than 30 dB, the proposed method outperforms all baseline schemes, which means the performance upper bound of the proposed method is the best.
\par Considering that both the proposed method and LISTA use layer-wise or block-wise training strategy, we compare their NMSE over iterations in noiseless scenario. Fig. 2(b) shows that the proposed method outperforms the LISTA method when using the same number of blocks in this scenario. 
\par We compare the performance of our proposed method with and without data consistency projection. Fig 2(c) indicates that data consistency brings a huge boost for performance. Without data consistency projection, the proposed method converges earlier and falls into a local solution. Under this circumstance, the generator is equivalent to AE and VAE with shortcut connection, which in turn implies that the performance of AE and VAE is not good for JADCE problem and data consistency projection makes the network learning ability stronger. Fig. 2(d) shows the performance difference of our method with and without training pilot matrix $\tilde{\bm{S}}$. We can observe that the optimization about pilot matrix can alleviate the influence of noise to improve the performance about 3 dB. 
\section{Conclusion}
In this paper, we proposed a novel model-free method to address the JADCE problem in IoT networks. We adopted U-net structure to build generator and used $\bm{S}^+\bm{Y}$ as input to generator. Inspired by deep residual learning and the properties of the Moore–Penrose pseudoinverse, we used $\bm{X}_1=\bm{S}^+\bm{Y}+(\bm{I}-\bm{S}^+\bm{S})U_1(\bm{S}^+\bm{Y})$ to ensure the data consistency. To reduce the impact of noise, we built a  two-layer fully-connected neural network to design pilot matrix. Simulation results showed that the data consistency projection and designing pilot matrix can improve the performance of the proposed method. In high SNR regimes, the proposed method achieves a better performance for the JADCE problem.

\bibliographystyle{IEEEbib}

\bibliography{refs}

\begin{thebibliography}{10}

\bibitem{massive_connectivity}
S.~Sharma and X.~Wang,
\newblock ``Toward massive machine type communications in ultra-dense cellular
  iot networks: Current issues and machine learning-assisted solutions,''
\newblock {\em IEEE Commmun. Surv. \& Tut.}, vol. 22, no. 1, pp. 426--471, Jan.
  2019.

\bibitem{Shi2022Algorithm}
Y.~Shi, H.~Choi, Y.~Shi, and Y.~Zhou,
\newblock ``Algorithm unrolling for massive access via deep neural networks
  with theoretical guarantee,''
\newblock {\em IEEE Trans. Wireless Commun.}, to appear.

\bibitem{Zou2021Learning}
Y.~Zou, Y.~Zhou, Y.~Shi, and X.~Chen,
\newblock ``Learning proximal operator methods for massive connectivity in
  {IoT} networks,''
\newblock in {\em Proc. IEEE Globecom}, Madrid Madrid, Spain, Dec. 2021.

\bibitem{grant_free}
L.~Liu, E.~Larsson, W.~Yu, P.~Popovski, C.~Stefanovic, and E.~de~Carvalho,
\newblock ``Sparse signal processing for grant-free massive connectivity: A
  future paradigm for random access protocols in the internet of things,''
\newblock {\em IEEE Signal Process. Mag.}, vol. 35, no. 5, pp. 88--99, 2018.

\bibitem{JADCE}
Z.~Chen, F.~Sohrabi, and W.~Yu,
\newblock ``Sparse activity detection for massive connectivity,''
\newblock {\em IEEE Trans. on Signal Process.}, vol. 66, no. 7, pp. 1890--1904,
  Jul. 2018.

\bibitem{Mostafa2019}
A.~Mostafa, Y.~Zhou, and V.~W.S. Wong,
\newblock ``Connection density maximization of narrowband iot systems with
  noma,''
\newblock {\em IEEE Trans. Wireless Commun.}, vol. 18, no. 10, pp. 4708--4722,
  Oct. 2019.

\bibitem{yuan2006model}
M.~Yuan and Y.~Lin,
\newblock ``Model selection and estimation in regression with grouped
  variables,''
\newblock {\em Journal of the Royal Statistical Society: Series B (Statistical
  Methodology)}, vol. 68, no. 1, pp. 49--67, 2006.

\bibitem{ista}
Z.~Qin, K.~Scheinberg, and D.~Goldfarb,
\newblock ``Efficient block-coordinate descent algorithms for the group
  lasso,''
\newblock {\em Math. Program. Comput.}, vol. 5, no. 2, pp. 143--169, 2013.

\bibitem{lista2}
Y.~{Shi}, S.~{Xia}, Y.~{Zhou}, and Y.~{Shi},
\newblock ``Sparse signal processing for massive device connectivity via deep
  learning,''
\newblock in {\em 2020 IEEE Int. Conf. on Commun. Workshops}, 2020, pp. 1--6.

\bibitem{CY}
Y.~Cui, S.~Li, and W.~Zhang,
\newblock ``Jointly sparse signal recovery and support recovery via deep
  learning with applications in mimo-based grant-free random access,''
\newblock {\em IEEE J. Sel. Areas Commun.}, vol. 39, no. 3, pp. 788--803, 2021.

\bibitem{VAE}
C.~Doersch,
\newblock ``Tutorial on variational autoencoders,''
\newblock {\em arXiv preprint arXiv:1606.05908}, 2016.

\bibitem{gan}
I.~Goodfellow, J.~Pouget-Abadie, M.~Mirza, B.~Xu, D.~Warde-Farley, S.~Ozair,
  A.~Courville, and Y.~Bengio,
\newblock ``Generative adversarial nets,''
\newblock {\em Advances in neural inf. process. syst.}, vol. 27, 2014.

\bibitem{AE}
S.~Li, W.~Zhang, Y.~Cui, H.~Cheng, and W.~Yu,
\newblock ``Joint design of measurement matrix and sparse support recovery
  method via deep auto-encoder,''
\newblock {\em IEEE Signal Proces. Lett.}, vol. 26, no. 12, pp. 1778--1782,
  Dec. 2019.

\bibitem{ganhigh}
E.~Balevi, A.~Doshi, A.~Jalal, A.~Dimakis, and J.~Andrews,
\newblock ``High dimensional channel estimation using deep generative
  networks,''
\newblock {\em IEEE J. Sel. Areas Commun.}, vol. 39, no. 1, pp. 18--30, Jan.
  2020.

\bibitem{unet}
O.~Ronneberger, P.~Fischer, and T.~Brox,
\newblock ``U-net: Convolutional networks for biomedical image segmentation,''
\newblock in {\em Int. Conf. on Med. Image Computing and Computer-Assisted
  Intervention}. Springer, 2015, pp. 234--241.

\bibitem{Unet1}
T.~Quan, T.~Nguyen-Duc, and W.~Jeong,
\newblock ``Compressed sensing {MRI} reconstruction using a generative
  adversarial network with a cyclic loss,''
\newblock {\em IEEE Trans. on Med. Imag.}, vol. 37, no. 6, pp. 1488--1497, Jun.
  2018.

\bibitem{Unet2}
M.~Mardani, E.~Gong, J.~Cheng, S.~Vasanawala, G.~Zaharchuk, M.~Alley,
  N.~Thakur, S.~Han, W.~Dally, J.~Pauly, et~al.,
\newblock ``Deep generative adversarial networks for compressed sensing
  automates mri,''
\newblock {\em arXiv preprint arXiv:1706.00051}, 2017.

\bibitem{LSGAN}
X.~Mao, Q.~Li, H.~Xie, R.~Lau, Z.~Wang, and S.~Paul Smolley,
\newblock ``Least squares generative adversarial networks,''
\newblock in {\em Proc. of IEEE Int. Conf. on Computer Vision}, 2017, pp.
  2794--2802.

\bibitem{wgan}
M.~Arjovsky, S.~Chintala, and L.~Bottou,
\newblock ``{W}asserstein generative adversarial networks,''
\newblock in {\em Proc. of the 34th International Conference on Machine
  Learning}, Aug. 2017, vol.~70, pp. 214--223.

\end{thebibliography}

\end{document}